\documentclass[12pt,onecolumn,draftclsnofoot]{IEEEtran}
\IEEEoverridecommandlockouts
\usepackage{cite}
\usepackage{amsmath,amssymb,amsfonts}
\usepackage{textcomp}
\usepackage{xcolor}
\usepackage{multirow}
\usepackage{graphicx}
\usepackage{textcomp}
\usepackage{float}
\usepackage{xcolor}
\usepackage{booktabs}
\usepackage{multirow}
\usepackage{subcaption}
\usepackage{algorithm, algpseudocode, graphicx}
\usepackage{hyperref}
\usepackage[utf8]{inputenc}
\def\BibTeX{{\rm B\kern-.05em{\sc i\kern-.025em b}\kern-.08em
    T\kern-.1667em\lower.7ex\hbox{E}\kern-.125emX}}
\begin{document}

\title{Trustworthy Efficient Communication for Distributed Learning using LQ-SGD Algorithm\\
\thanks{}
}

\author{Hongyang Li, Lincen Bai, Caesar Wu, Mohammed Chadli, Said Mammar, Pascal Bouvry}

\maketitle

\begingroup
\renewcommand\thefootnote{}\footnotetext{
\textsuperscript{1}University of Luxembourg, Luxembourg. \{hongyang.li, caesar.wu, pascal.bouvry\}@uni.lu

\textsuperscript{2}University of Paris-Saclay, France. \{lincen.bai, said.mammar, mohammed.chadli\}@univ-evry.fr

\textsuperscript{}.
}
\endgroup

\begin{abstract}

We propose LQ-SGD (Low-Rank Quantized Stochastic Gradient Descent), an efficient communication gradient compression algorithm designed for distributed training. LQ-SGD further develops on the basis of PowerSGD by incorporating the low-rank approximation and log-quantization techniques, which drastically reduce the communication overhead, while still ensuring the convergence speed of training and model accuracy. In addition, LQ-SGD and other compression-based methods show stronger resistance to gradient inversion than traditional SGD, providing a more robust and efficient optimization path for distributed learning systems.

\end{abstract}

\begin{IEEEkeywords}
distributed Learning, gradient compression, low-rank approximation, logarithmic quantization, communication efficiency, gradient inversion, trustworthiness, stochastic gradient descent
\end{IEEEkeywords}

\vspace{-3mm}
\section{Introduction}

With the rapid development of learning models, distributed training has become a fundamental approach to improving model performance and scalability. However, these distributed training systems typically rely on numerous compute nodes working collaboratively, where the synchronization of model parameters and gradients introduces significant communication overhead. As model sizes continue to increase and the number of nodes expands, communication overhead has become the primary bottleneck. In large-scale models, communication time even surpasses computation time, severely degrading the overall efficiency of the training process~\cite{anthony2024demystifying, goyal2017accurate, shallue2018measuring, lin2018dgc, chen2023communication}. In addition to efficiency concerns, the trustworthiness and data privacy risks associated with distributed learning are also becoming increasingly critical\cite{li2025lightweight, geiping2020inverting, li2023adaptive}. Frequent exchanges of gradient information may inadvertently expose sensitive training data, making distributed learning systems vulnerable to privacy attacks.

To reduce communication bottlenecks, a variety of gradient compression techniques have been proposed~\cite{li2022communication, seide20141bit, alistarh2017qsgd, lin2018dgc, stich2018sparsified, vogels2019powersgd}. Among them, PowerSGD leverages low-rank approximation to significantly reduce communication overhead while maintaining competitive model performance, and has become one of the state-of-the-art solutions for distributed optimization. However, existing methods still face limitations in terms of compression efficiency. 

To address these issues, we propose a novel gradient compression algorithm for distributed learning systems, namely \textbf{LQ-SGD}. LQ-SGD builds upon the PowerSGD \cite{vogels2019powersgd} framework by introducing a logarithmic gradient quantization mechanism, combining low-rank approximation with quantization strategies. This design further reduces communication overhead while effectively maintaining convergence speed and model accuracy. Here:

\begin{enumerate} 
 \item We propose LQ-SGD, an innovative algorithm to further reduce the communication cost by incorporating log-quantization techniques into PowerSGD.

 \item We evaluate the convergence performance and communication efficiency of LQ-SGD on multiple standard datasets (e.g., MNIST, CIFAR-10/100).

\item We explore the impact of gradient compression on model trustworthiness and that possesses stronger robustness in defending against gradient inversion attacks.
\end{enumerate}

\vspace{-3mm}

\section{Related Work}

\subsection{Communication Bottleneck in Distributed Training}

In large-scale systems, communication time often surpasses computation time, becoming the primary bottleneck that limits overall training efficiency~\cite{goyal2017accurate}. For example, in training a 1.3-billion parameter model, communication accounts for 50\% to 90\% of the total training time as the system scales~\cite{anthony2024demystifying}. This severe imbalance between computation and communication makes it difficult to fully utilize distributed computational resources. Reducing communication costs is therefore crucial for enabling efficient and scalable distributed learning. This motivates the development of communication-efficient techniques.
\vspace{-3mm}

\subsection{Gradient Compression Techniques}

To reduce communication costs in distributed learning, three mainstream gradient compression techniques have been widely adopted: quantization, sparsification, and low-rank approximation. Quantization methods, such as 1-bit SGD~\cite{seide20141bit} and QSGD~\cite{alistarh2017qsgd}, reduce the precision of gradients to lower the communication bandwidth. Sparsification techniques, including Top-K selection~\cite{aji2017sparse} and Deep Gradient Compression (DGC)~\cite{lin2018dgc}, transmit only the some significant gradients. Low-rank approximation methods, such as PowerSGD~\cite{vogels2019powersgd}, compress gradients by approximating them with low-rank matrices, achieving substantial communication reduction in structured layers.

    \vspace{-3mm}

\section{Preliminaries}

\subsection{Distributed SGD Basics}

Distributed Stochastic Gradient Descent (Distributed SGD) is a widely used framework for accelerating large-scale model training. In this paradigm, the global training dataset is partitioned into subsets, which are distributed across $N$ compute nodes (or workers). At each iteration $t$, each worker $i$ independently computes its local gradient $\mathbf{g}_t^{(i)}$ on its data shard. These local gradients are then aggregated across all workers—typically via an \textit{All-Reduce} operation—to obtain the global gradient $\mathbf{g}_t$:
\begin{equation}
    \mathbf{g}_t = \frac{1}{N} \sum_{i=1}^{N} \mathbf{g}_t^{(i)}.
\end{equation}

After aggregation, the model parameters $\mathbf{w}_t$ are updated synchronously using the global gradient:
\begin{equation}
\vspace{-2mm}
    \mathbf{w}_{t+1} = \mathbf{w}_t - \eta \mathbf{g}_t,
\end{equation}
where $\eta$ is the learning rate.

\vspace{-2mm}
\subsection{PowerSGD}
Unlike quantization or sparsification methods that compress gradients element-wise, PowerSGD~\cite{vogels2019powersgd} exploits the low-rank structure of gradient matrices—especially in large neural networks—to achieve significant communication reduction.

In typical deep learning models, such as convolutional and fully-connected layers, the gradients of weight matrices often have a high degree of redundancy. PowerSGD approximates these high-dimensional gradients using low-rank matrix factorization before communication. Specifically, given a gradient matrix $G \in \mathbb{R}^{m \times n}$, PowerSGD approximates $G$ as a product of two low-rank matrices:
\begin{equation}
    G \approx PQ^\top
\end{equation}
where $P \in \mathbb{R}^{m \times r}$ and $Q \in \mathbb{R}^{n \times r}$, and $r \ll \min(m, n)$ is the rank of the approximation.


After the step of aggregation in  distributed learning , $P$ and $Q$ are used to reconstruct an approximation of $G$. Since $P$ and $Q$ are much smaller than $G$ (when $r$ is small), this approach drastically reduces the amount of data transmitted during synchronization. PowerSGD achieves compression rates proportional to $r(m + n) / (mn)$, making it highly scalable for layers with large weight matrices. 

\subsection{Model Trustworthiness}

In distributed and federated learning, the frequent exchange of gradients exposes models to significant privacy risks. One of the most critical threats is the \textbf{Gradient Inversion Attack} (GIA), which aims to reconstruct the original training data from shared gradients.

Given the gradient \( \mathbf{g}_t \) received by the server or other participants, the attacker reconstructs dummy inputs \( \hat{\mathbf{x}} \) by minimizing the cosine similarity loss between the gradient \( \mathbf{g}_t \), along with a regularization term~\cite{geiping2020inverting}:
\begin{equation}
    \hat{\mathbf{x}} = \arg\min_{\mathbf{x}} \left[ 1 - \frac{\left\langle \nabla_{\mathbf{w}} \mathcal{L}(f(\mathbf{x}; \mathbf{w}), y), \mathbf{g}_t \right\rangle}{\left\| \nabla_{\mathbf{w}} \mathcal{L}(f(\mathbf{x}; \mathbf{w}), y) \right\|_2 \left\| \mathbf{g}_t \right\|_2} \; + \; R(\mathbf{x}) \right],
\end{equation}
where \( \mathcal{L} \) is the loss function, \( f(\cdot; \mathbf{w}) \) is the model, \( y \) is an optional label (depending on the attack setting), and \( R(\mathbf{x}) \) is a regularization term, such as total variation (TV), to improve reconstruction quality.

\vspace{-3mm}
\section{Proposed Method}

By incorporating log-quantization techniques, we present here the proposed LQ-SGD algorithm, which is an enhanced version of PowerSGD~\cite{vogels2019powersgd}. 

While PowerSGD effectively reduces the communication cost by compressing the gradient matrix \( \mathbf{G}_t \in \mathbb{R}^{n \times m} \) into two low-rank factor matrices \( \mathbf{P}_t \in \mathbb{R}^{n \times r} \) and \( \mathbf{Q}_t \in \mathbb{R}^{m \times r} \), the size of these factor matrices remains substantial, especially when a higher approximation rank \( r \) is needed to maintain convergence. To further reduce the communication overhead, we introduce a logarithmic quantization strategy. This method achieves efficient compression with little impact on the accuracy of numerical representations by assigning higher precision to smaller values. Meanwhile, LQ-SGD, like PowerSGD, incorporates an error feedback mechanism and a hot-start initialization strategy, thus guaranteeing stable convergence of the algorithm even when high-intensity quantization is employed. The algorithm procedure of LQ-SGD has be summarized in \textbf{Algorithm~\ref{alg:LQ-SGD}}.

\subsection{Logarithmic Quantization of \( \mathbf{P}_t \) and \( \mathbf{Q}_t \)}
 
Since heavy-tailed behavior is commonly observed in practical scenarios \cite{hodgkinson2025models}, it is essential to design quantization schemes that account for such characteristics. In the presence of heavy-tailed or power-law distributed gradients, nonuniform quantization methods are particularly effective. These schemes allocate higher precision to smaller gradient values—where the majority of informative signals are concentrated—while more aggressively compressing large outliers, thereby reducing redundancy and improving overall efficiency. 

We adopt the following quantization function:
\begin{equation}
    q(x) = \text{sign}(x) \cdot \frac{\log(1 + \alpha |x|)}{\log(1 + \alpha)}, \quad \alpha > 0,
\end{equation}
where \( \alpha \) controls the curvature of the logarithmic mapping. This design prioritizes accuracy near the zero value, since these small amplitudes form a major part of the typical gradient distribution ~\cite{alistarh2017qsgd, lin2018dgc}.

The matrices after quantizing \( \mathbf{P}_{\text{quant}} \) and \( \mathbf{Q}_{\text{quant}} \) are sent across workers via the All-Reduce operation. Upon reception, each worker will dequantize to reconstruct the original information:
\begin{equation}
    x = \text{sign}(q(x)) \cdot \frac{(1 + \alpha)^{|q(x)|} - 1}{\alpha}.
\end{equation}

In practical deployments, we adopt a separable symbol encoding scheme. The normalized quantized values \( |q(x)| \in [0,1] \) are mapped to a discrete set of \( 2^b \) uniformly spaced bins, determined by a \( b \)-bit representation. Under this design, each quantized scalar requires only \( b \) bits for transmission; in our experiments, we typically set \( b = 8 \). To implement this efficiently, we discretize the continuous log-based function via precomputed quantization levels and nearest-neighbor matching. This strategy retains the precision structure of logarithmic mapping and supports importance-aware compression~\cite{ramezaniprovably}.

\subsection{Error Feedback Mechanism}

Through the error feedback mechanism, LQ-SGD compensates the signal distortion caused by low-rank approximation and quantization to a certain extent. After the gradient approximation reconstruction is completed :

\begin{equation}
    \hat{\mathbf{G}}_t = \mathbf{P}_t \mathbf{Q}_t^\top,
\end{equation}

\vspace{-1.5ex}

the residual error is computed:
\begin{equation}
    \mathbf{E}_t = \mathbf{G}_t' - \hat{\mathbf{G}}_t.
\end{equation}

\vspace{-1.5ex}

This error is accumulated and added to the gradient in the next iteration:
\begin{equation}
    \mathbf{G}_{t+1}' = \mathbf{G}_{t+1} + \mathbf{E}_t.
\end{equation}

This technique is often referred to as an error feedback mechanism , such as EF-SGD~\cite{karimireddy2019error}, which establishes the corresponding theoretical framework. Some low-rank compression methods in recent years, such as PowerSGD~\cite{vogels2019powersgd}, also introduce error feedback in their design to compensate for the errors caused by the low-rank approximation.

In LQ-SGD, we adopt this mechanism to ensure that the information lost due to low-rank factorization and logarithmic quantization is gradually recovered, preserving the convergence properties.

\begin{algorithm}[H]
\caption{Low-Rank Quantized Stochastic Gradient Descent (LQ-SGD)}
\label{alg:LQ-SGD}
\begin{algorithmic}[1]

\Require Gradient matrix \( \mathbf{G}_t \in \mathbb{R}^{n \times m} \), rank \( r \), quantization bits \( b_p, b_q \), logarithmic scale \( \alpha \), learning rate \( \eta \)
\Ensure Updated model parameters \( \mathbf{w}_{t+1} \)

\State Initialize error matrix \( \mathbf{E}_0 \gets \mathbf{0} \)
\State Initialize \( \mathbf{Q}_0 \sim \mathcal{N}(0, 1) \)

\For{each iteration \( t = 1, 2, \dots \)}

    \State \( \mathbf{G}_t' \gets \mathbf{G}_t + \mathbf{E}_{t-1} \) \Comment{Error feedback compensation}

    \If{\( t > 1 \)}
        \State \( \mathbf{Q}_t \gets \mathbf{Q}_{t-1} \) \Comment{Warm-start from previous \( \mathbf{Q} \)}
    \Else
        \State \( \mathbf{Q}_t \sim \mathcal{N}(0, 1) \) \Comment{Random initialization}
    \EndIf

    \State \( \mathbf{P}_t \gets \mathbf{G}_t' \mathbf{Q}_t \)
    \State \( \mathbf{P}_t \gets \text{Orthonormalize}(\mathbf{P}_t) \) \Comment{Power iteration step}

    \State \( \mathbf{P}_{\text{quant}} \gets \text{LogQuantize}(\mathbf{P}_t, b_p, \alpha) \)
    \State All-Reduce \( \mathbf{P}_{\text{quant}} \)
    \State \( \mathbf{P}_t \gets \text{LogDequantize}(\mathbf{P}_{\text{quant}}, b_p, \alpha) \)

    \State \( \mathbf{Q}_t \gets \mathbf{G}_t'^\top \mathbf{P}_t \)
    \State \( \mathbf{Q}_{\text{quant}} \gets \text{LogQuantize}(\mathbf{Q}_t, b_q, \alpha) \)
    \State All-Reduce \( \mathbf{Q}_{\text{quant}} \)
    \State \( \mathbf{Q}_t \gets \text{LogDequantize}(\mathbf{Q}_{\text{quant}}, b_q, \alpha) \)

    \State \( \hat{\mathbf{G}}_t \gets \mathbf{P}_t \mathbf{Q}_t^\top \) \Comment{Reconstruct gradient}
    \State \( \mathbf{E}_t \gets \mathbf{G}_t' - \hat{\mathbf{G}}_t \) \Comment{Update error feedback}

    \State \( \mathbf{w}_{t+1} \gets \mathbf{w}_t - \eta \hat{\mathbf{G}}_t \) \Comment{Model update}

\EndFor

\end{algorithmic}
\end{algorithm}

\subsection{Communication and Computational Complexity}

Compared to PowerSGD, which requires transmitting \( r(n + m) \) full-precision floating-point values (typically 32 bits each) per iteration, LQ-SGD reduces the communication cost to \( r(n + m) \times b \) bits per iteration, where \( b \) is the number of quantization bits used in \( \mathbf{P}_{\text{quant}} \) and \( \mathbf{Q}_{\text{quant}} \). This achieves a compression ratio of \( 32 / b \) relative to PowerSGD.

The extra cost of quantization / de-quantization in LQ-SGD is only \(\mathcal{O}\!\bigl(r(n+m)\bigr)\), whereas the two
matrix–matrix products in PowerSGD dominate at \(\mathcal{O}(nmr)\), the added computation is practically negligible on modern GPUs.

\section{Experimental Results}

In this section, we evaluate the performance and trustworthiness of our proposed LQ-SGD algorithm in distributed training tasks.
\vspace{-3mm
}
\subsection{Experimental Setup}

The experiments are conducted on a distributed cluster comprising 5 worker nodes, each equipped with an NVIDIA RTX 4090 GPU, and 1 central node responsible for gradient aggregation. The system follows a parameter server-like architecture to simulate realistic distributed training conditions.

We evaluate the proposed \textbf{LQ-SGD} algorithm against three baselines: Original SGD, PowerSGD~\cite{vogels2019powersgd}, and TopK-SGD~\cite{shi2019distributed}. For TopK-SGD, the sparsity ratio is adjusted to achieve compression rates comparable to those of the other methods. Experiments are performed on three widely used image classification datasets: CIFAR-10, CIFAR-100, and MNIST. For the model architecture, we employ ResNet-18~\cite{he2016deep}, which offers a good balance between model complexity and computational efficiency in distributed environments. All algorithms are implemented within the same distributed training pipeline to ensure a fair and consistent comparison.

The evaluation covers the following metrics:
\begin{itemize}
    \item \textbf{Accuracy}: Top-1 test accuracy on the validation set.
    \item \textbf{Communication Cost}: The size (in MB) of gradient data transmitted per-epoch.
    \item \textbf{Computation Time}: The per-epoch computation time measured, excluding communication time (in seconds).
    \item \textbf{Structural Similarity Index Measure (SSIM)}: To evaluate privacy leakage resistance under GIA. Lower SSIM scores indicate better protection against data reconstruction from shared gradients.
\end{itemize}

\subsection{Performance Evaluation: Communication Cost, Computation Time, and Accuracy}
\begin{table}[h]

\vspace{-3mm}
    \centering
    \caption{Performance Comparison on CIFAR-10}
    \begin{tabular}{l|c|c|c}
        \toprule
        \textbf{Method} & \textbf{Accuracy} & \textbf{Size (x1)} & \textbf{Computing Time} \\
        \midrule
        Original SGD         & 0.9432 & 3325 (x1108.3) & 2.2937 \\
        Power SGD (Rank 1)   & 0.9451 & 14 (x4.7)      & 2.3359 \\
        TopK SGD$^\dagger$   & 0.8821 & 14 (x4.7)      & 3.6173 \\
        \textbf{LQ-SGD (Rank 1)} & \textbf{0.9290} & \textbf{3 (x1.0)} & 2.5714 \\
        \bottomrule
    \end{tabular}
    \label{tab:cifar10_level1}
\end{table}
\vspace{-2mm}

\begin{table}[h]
    \centering
    \caption{Performance Comparison on CIFAR-100}
    \begin{tabular}{l|c|c|c}
        \toprule
        \textbf{Method} & \textbf{Accuracy} & \textbf{Size (x1)} & \textbf{Computing Time} \\
        \midrule
        Original SGD         & 0.7445 & 3339 (x1113.0) & 2.2882 \\
        Power SGD (Rank 1)   & 0.7404 & 14 (x4.7)      & 2.1588 \\
        TopK SGD$^\dagger$   & 0.6070 & 14 (x4.7)      & 3.5946 \\
        \textbf{LQ-SGD (Rank 1)} & \textbf{0.7181} & \textbf{3 (x1.0)} & 2.5631 \\
        \bottomrule
    \end{tabular}
  
    \label{tab:cifar100_level1}
\end{table}

\begin{table}[h]
    \centering
    \caption{Performance Comparison on MNIST}
    \begin{tabular}{l|c|c|c}
        \toprule
        \textbf{Method} & \textbf{Accuracy} & \textbf{Size (x1)} & \textbf{Computing Time} \\
        \midrule
        Original SGD         & 0.9940 & 3964 (x991.0)  & 2.4909 \\
        Power SGD (Rank 1)   & 0.9929 & 16 (x4.0)      & 2.3617 \\
        TopK SGD$^\dagger$   & 0.9940 & 16 (x4.0)      & 3.9826 \\
        \textbf{LQ-SGD (Rank 1)} & \textbf{0.9939} & \textbf{4 (x1.0)} & 2.8442 \\
        \bottomrule
    \end{tabular}
    \caption*{\footnotesize $^\dagger$ TopK SGD achieves an effective compression ratio aligned with Power SGD (Rank 1) and LQ-SGD (Rank 1) in this experiment.}
    \label{tab:mnist_level1}

    \vspace{-3mm}
\end{table}

Table~\ref{tab:cifar10_level1}, ~\ref{tab:cifar100_level1} and  ~\ref{tab:mnist_level1} summarizes the accuracy, communication cost, and computation time of different algorithms on CIFAR-10, CIFAR-100, and MNIST at compression rank 1. LQ-SGD achieves substantial communication cost reduction compared to both PowerSGD and TopK-SGD. For example, on CIFAR-10, LQ-SGD reduces the communication volume by over 75\% relative to PowerSGD (3 MB vs. 14 MB), and by more than 99.9\% compared to uncompressed SGD (3 MB vs. 3325 MB). Similar reductions are observed on CIFAR-100 and MNIST, demonstrating LQ-SGD's ability to significantly lower bandwidth requirements. For comparison, TopK-SGD uses a sparsification strategy by transmitting only a fixed percentage of the largest gradient entries, achieving an effective compression ratio similar to PowerSGD at rank 1 (approximately 4.7$\times$ compression in our experiments). Despite its aggressive compression, LQ-SGD maintains high accuracy. On CIFAR-10, it achieves 92.9\% Top-1 accuracy, only slightly lower than PowerSGD (94.5\%) and outperforming TopK-SGD (88.2\%). On CIFAR-100 and MNIST, LQ-SGD shows similarly competitive accuracy while offering greater communication efficiency. LQ-SGD introduces only a marginal increase in computation time compared to PowerSGD, with the overhead from logarithmic quantization being minimal, and maintains reasonable per-epoch computation times across all datasets, remaining within 2.5 to 2.8 seconds per epoch on CIFAR-10, CIFAR-100, and MNIST.

\begin{figure*}[t]
    \centering
    \begin{subfigure}{0.4\textwidth}
        \centering
        \includegraphics[width=0.8\textwidth]{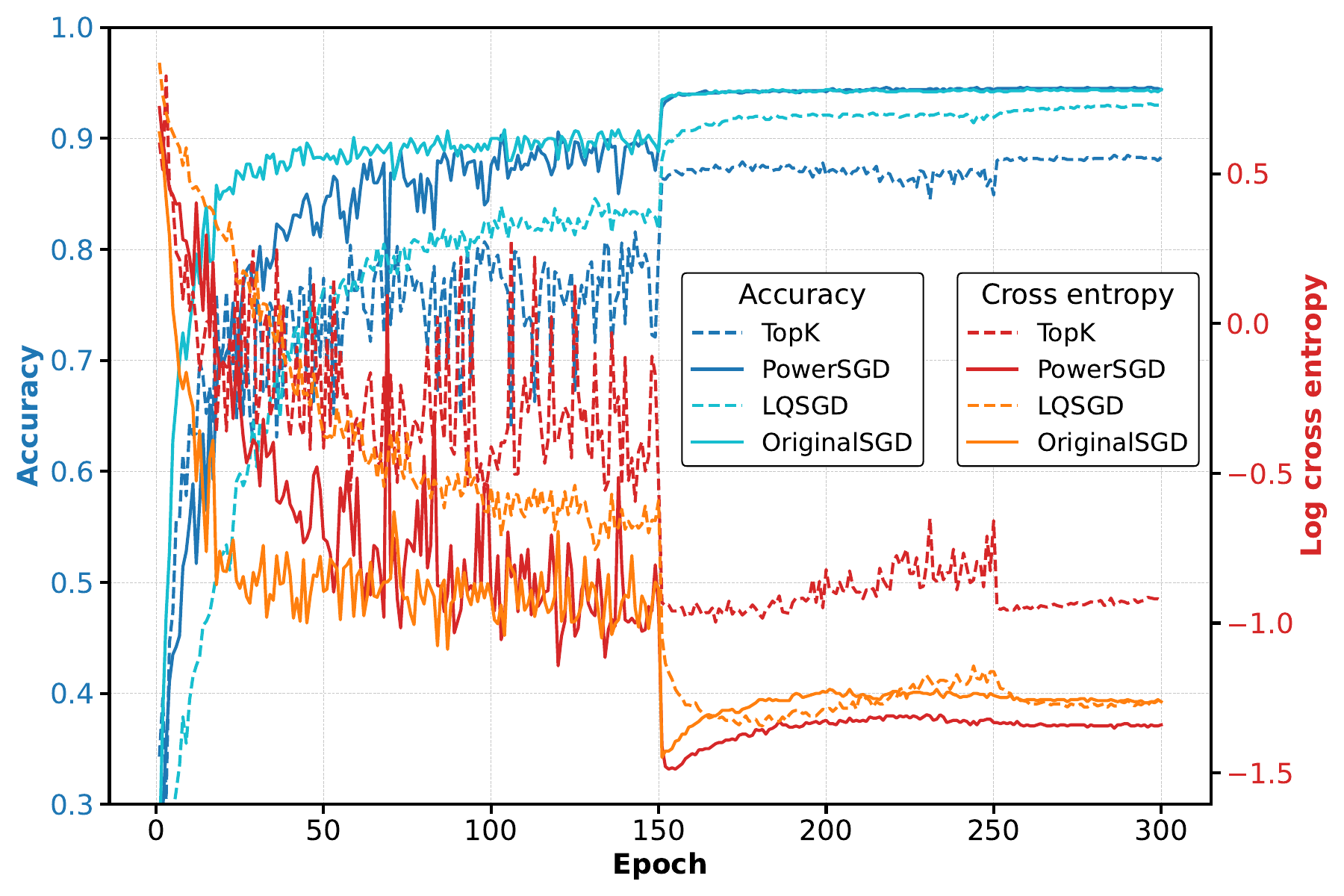}
        \caption{CIFAR-10, Rank 1}
        \label{fig:cifar10_rank1}
    \end{subfigure}
    \hfill
    \begin{subfigure}{0.4\textwidth}
        \centering
        \includegraphics[width=0.8\textwidth]{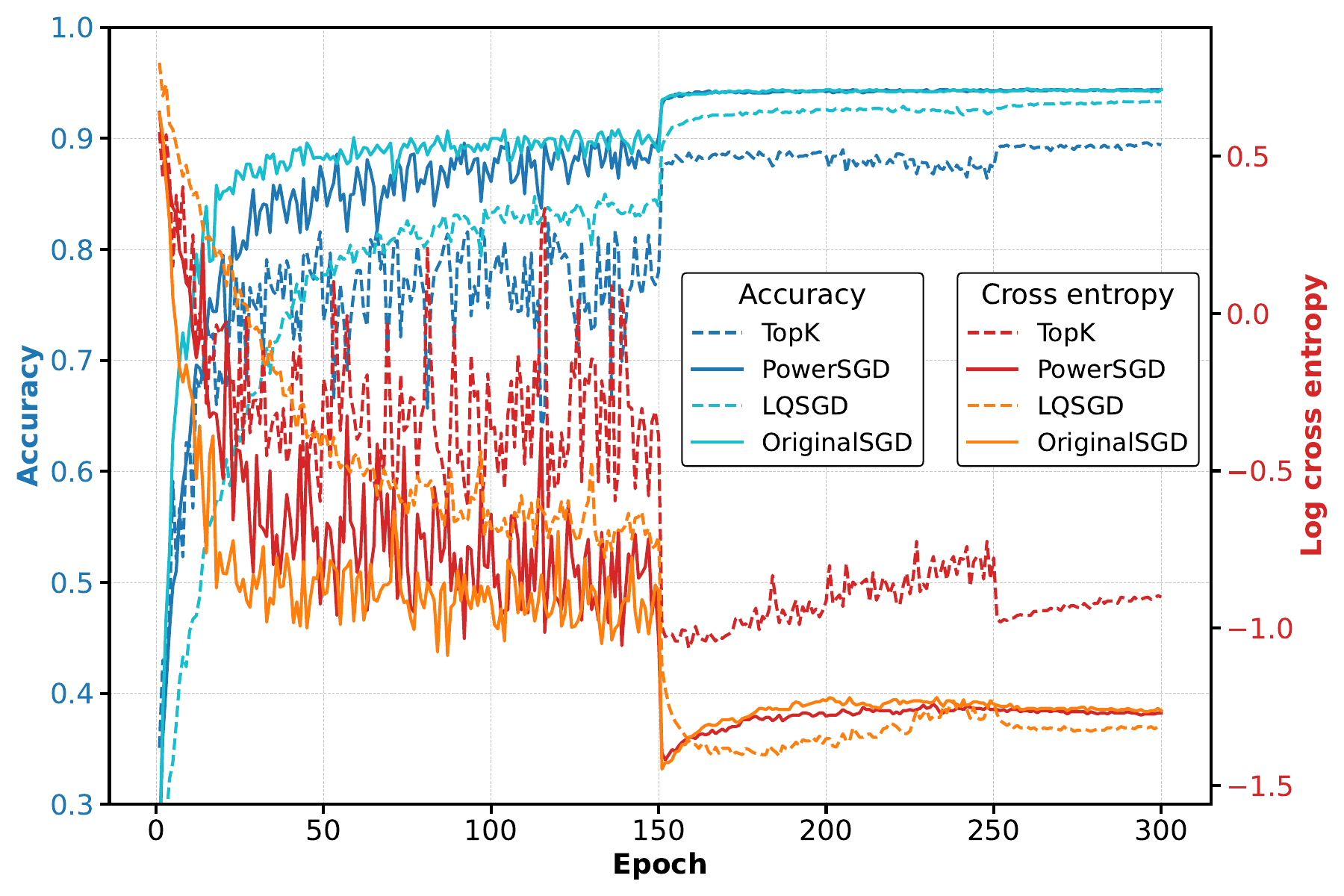}
        \caption{CIFAR-10, Rank 2}
        \label{fig:cifar10_rank2}
    \end{subfigure}
    \hfill
    \begin{subfigure}{0.4\textwidth}
        \centering
        \includegraphics[width=0.8\textwidth]{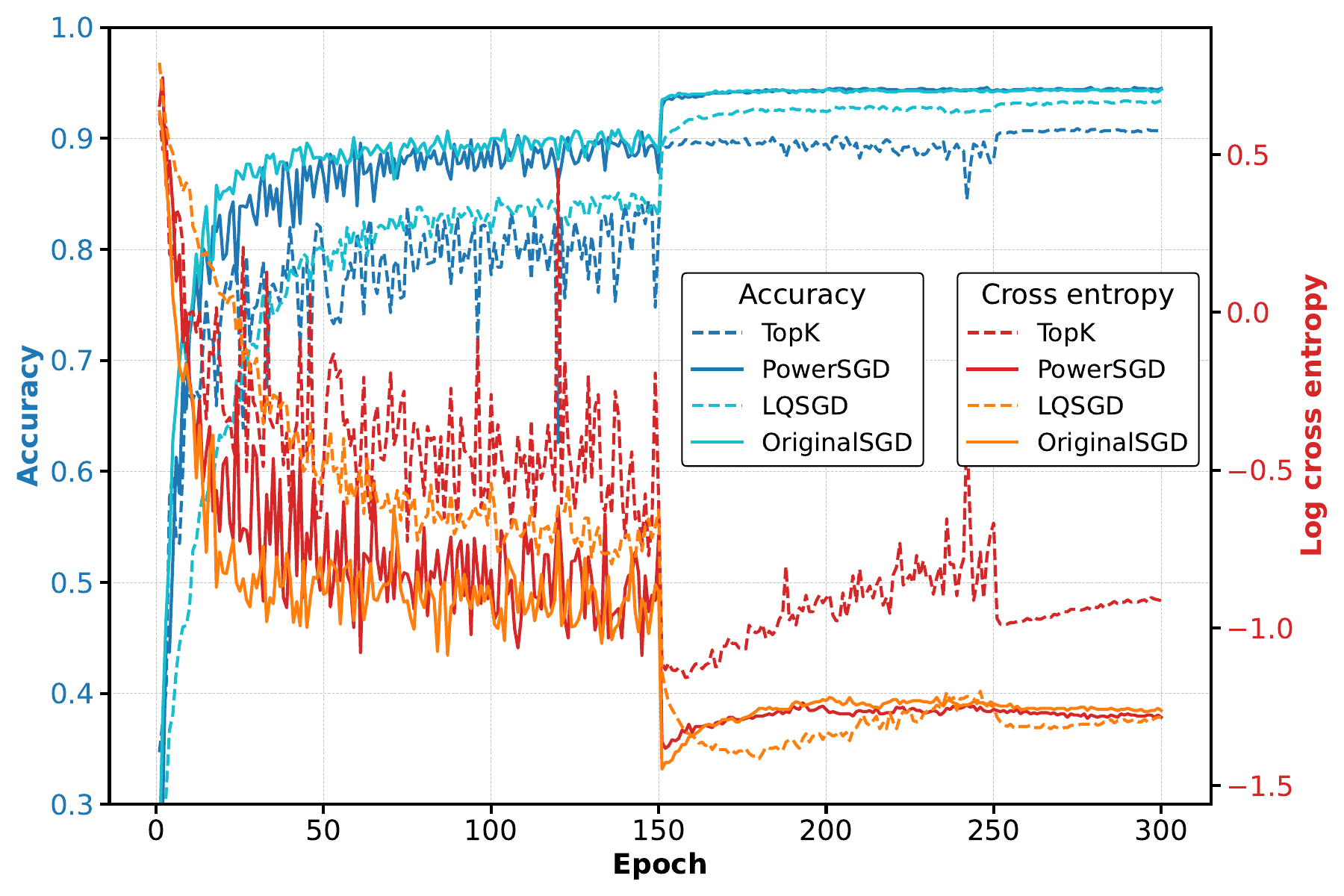}
        \caption{CIFAR-10, Rank 4}
        \label{fig:cifar10_rank4}
    \end{subfigure}
    \hfill
    \begin{subfigure}{0.4\textwidth}
        \centering
        \includegraphics[width=0.8\textwidth]{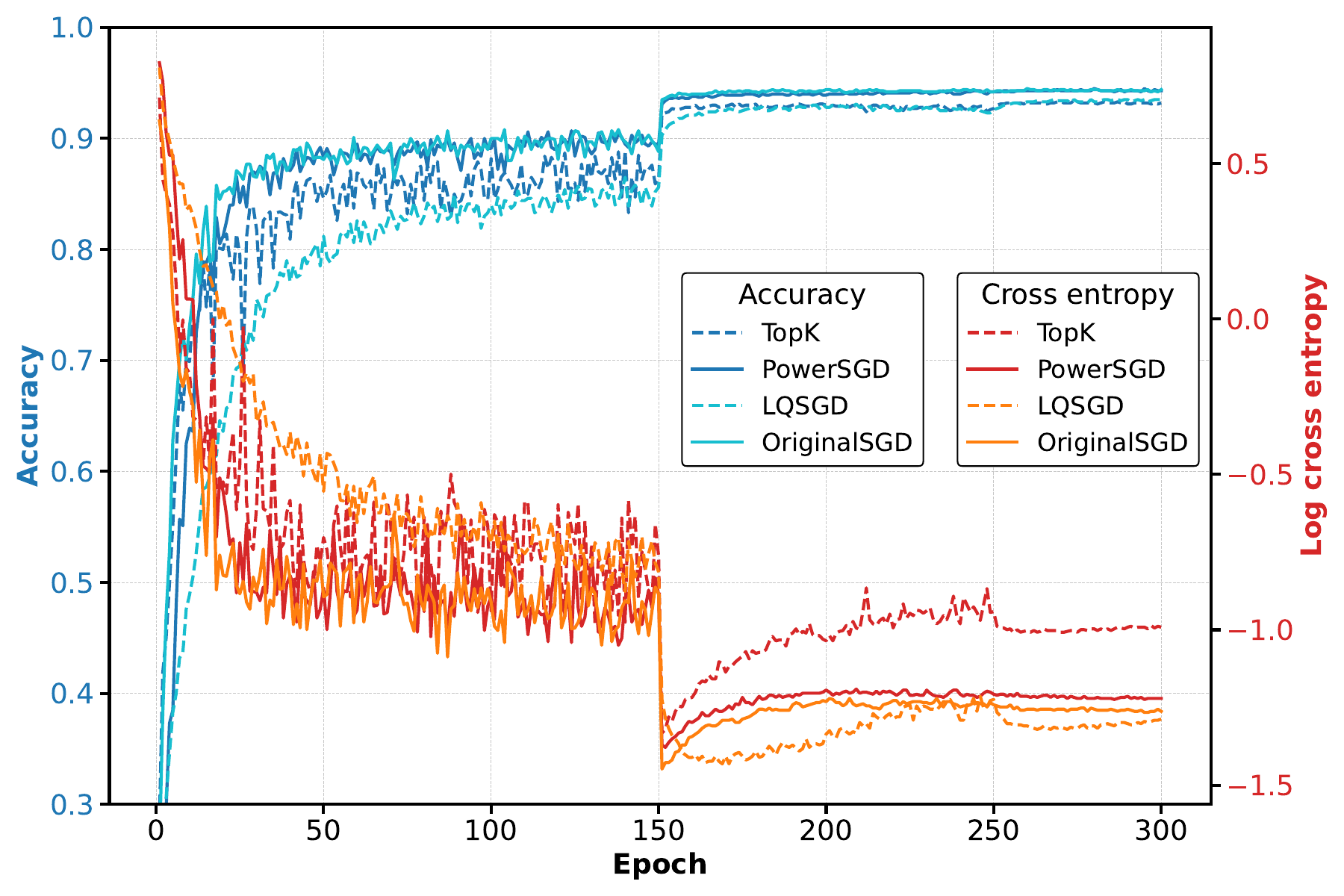}
        \caption{CIFAR-10, Rank 30}
        \label{fig:cifar10_rank30}
    \end{subfigure}
    \hfill
    \begin{subfigure}{0.4\textwidth}
        \centering
        \includegraphics[width=0.8\textwidth]{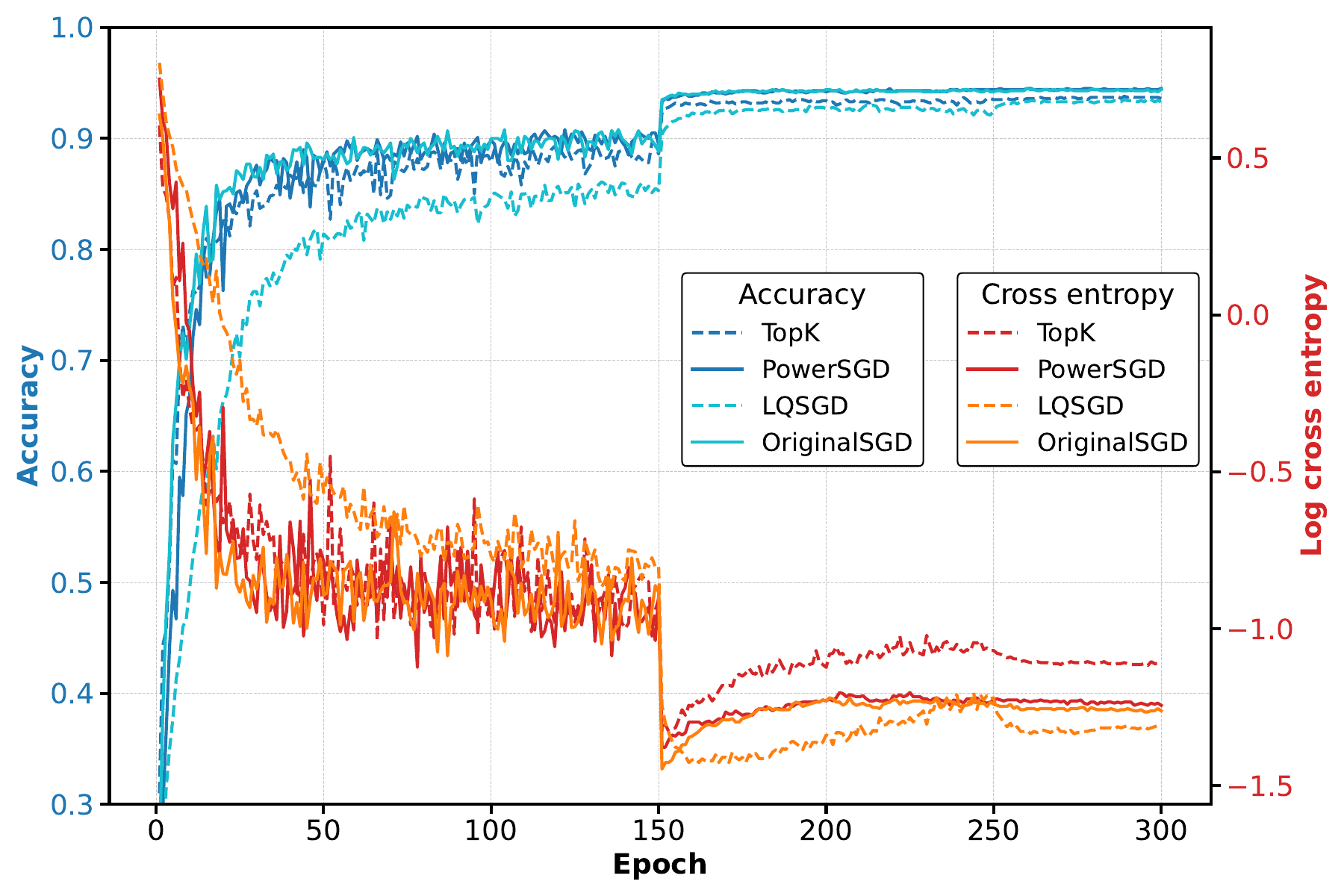}
        \caption{CIFAR-10, Rank 50}
        \label{fig:cifar10_rank50}
    \end{subfigure}
    
    \caption{Results for CIFAR-10 dataset with different compression ranks.}

    \label{fig:cifar10_selected_results}
\end{figure*}

\begin{figure*}[t]
    \centering
    \begin{subfigure}{0.4\textwidth}
        \centering
        \includegraphics[width=0.8\textwidth]{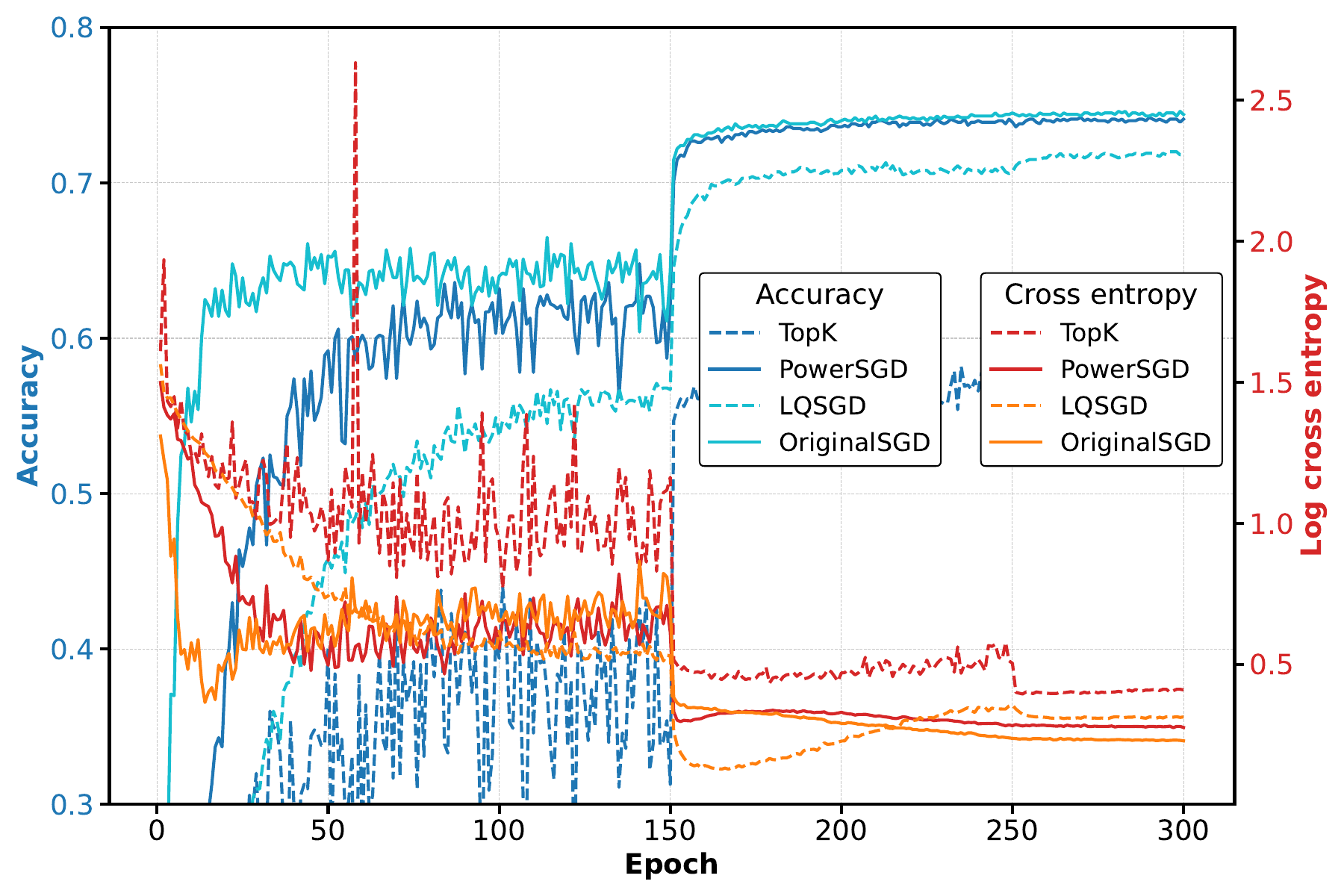}
        \caption{CIFAR-100, Rank 1}
        \label{fig:cifar100_rank1}
    \end{subfigure}
    \hfill
    \begin{subfigure}{0.4\textwidth}
        \centering
        \includegraphics[width=0.8\textwidth]{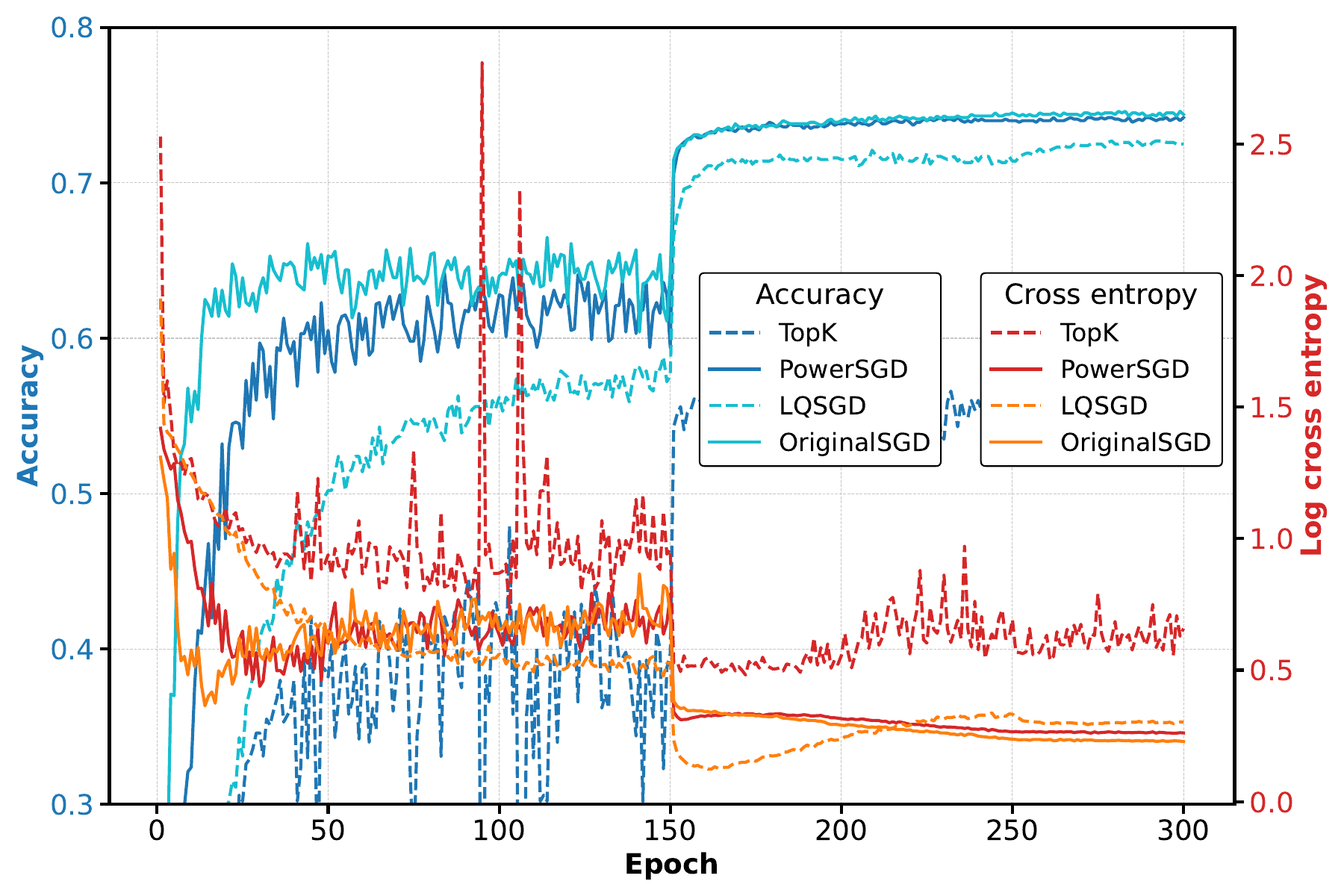}
        \caption{CIFAR-100, Rank 2}
        \label{fig:cifar100_rank2}
    \end{subfigure}
    \hfill
    \begin{subfigure}{0.4\textwidth}
        \centering
        \includegraphics[width=0.8\textwidth]{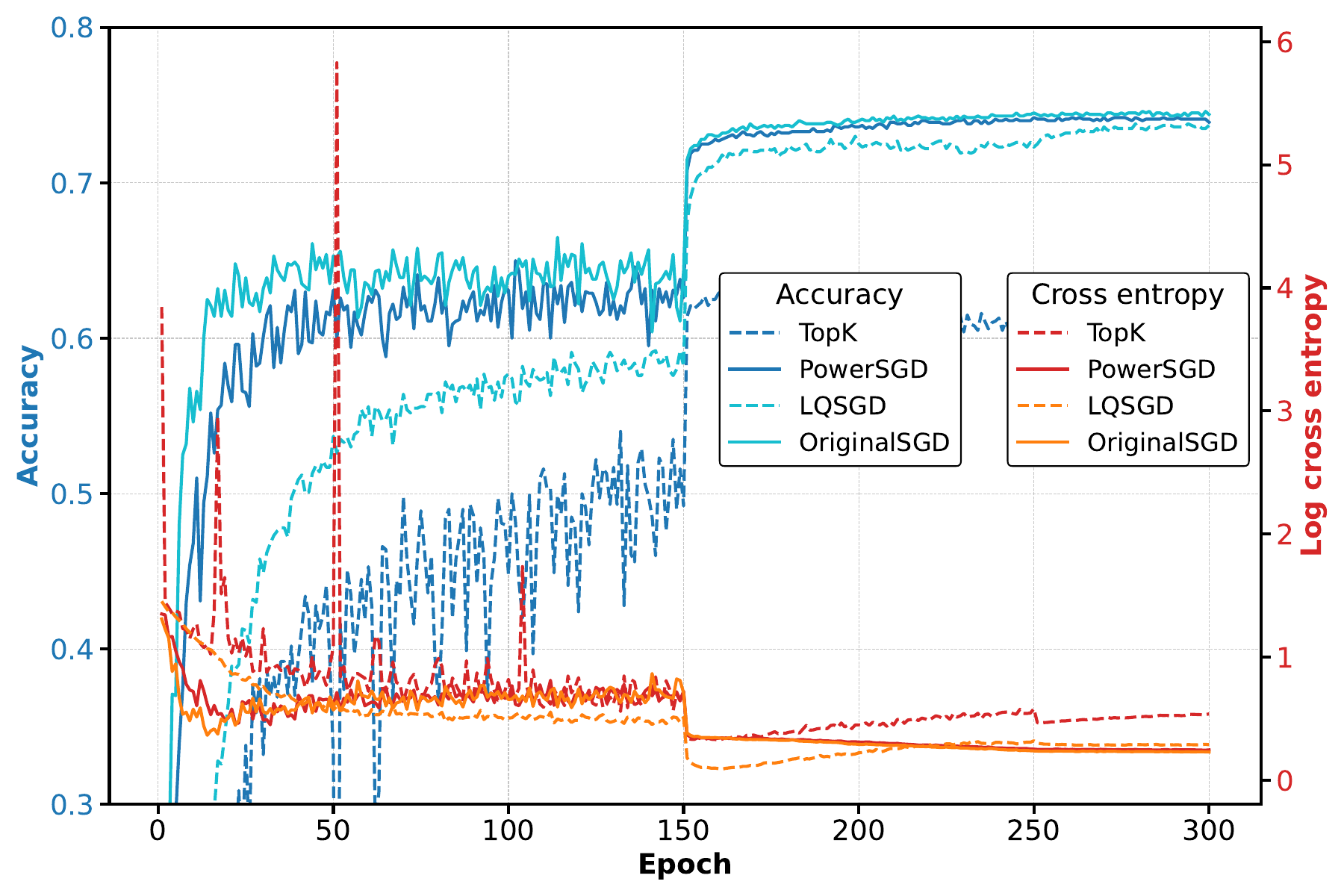}
        \caption{CIFAR-100, Rank 4}
        \label{fig:cifar100_rank4}
    \end{subfigure}
    \hfill
    \begin{subfigure}{0.4\textwidth}
        \centering
        \includegraphics[width=0.8\textwidth]{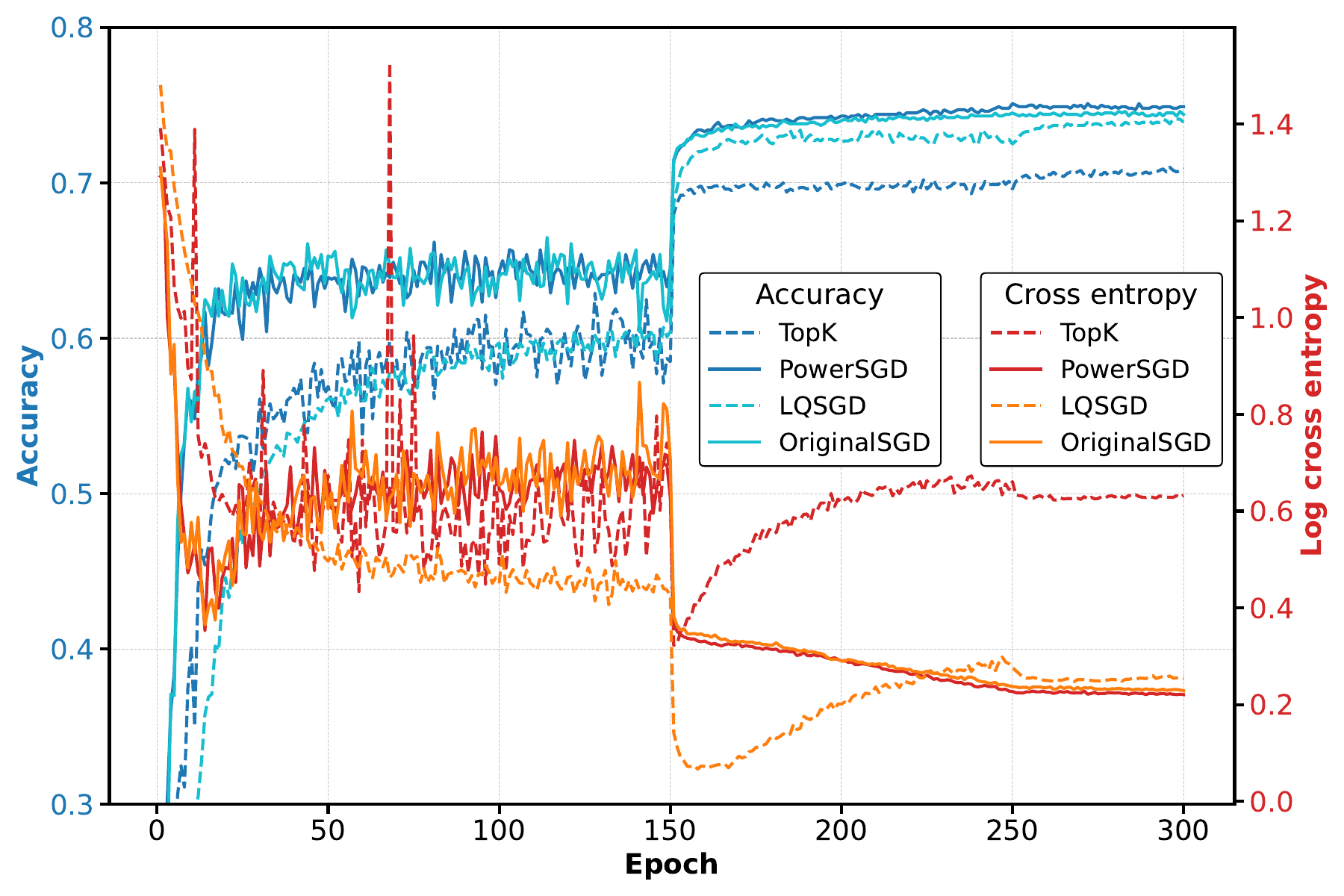}
        \caption{CIFAR-100, Rank 30}
        \label{fig:cifar100_rank30}
    \end{subfigure}
    \hfill
    \begin{subfigure}{0.4\textwidth}
        \centering
        \includegraphics[width=0.8\textwidth]{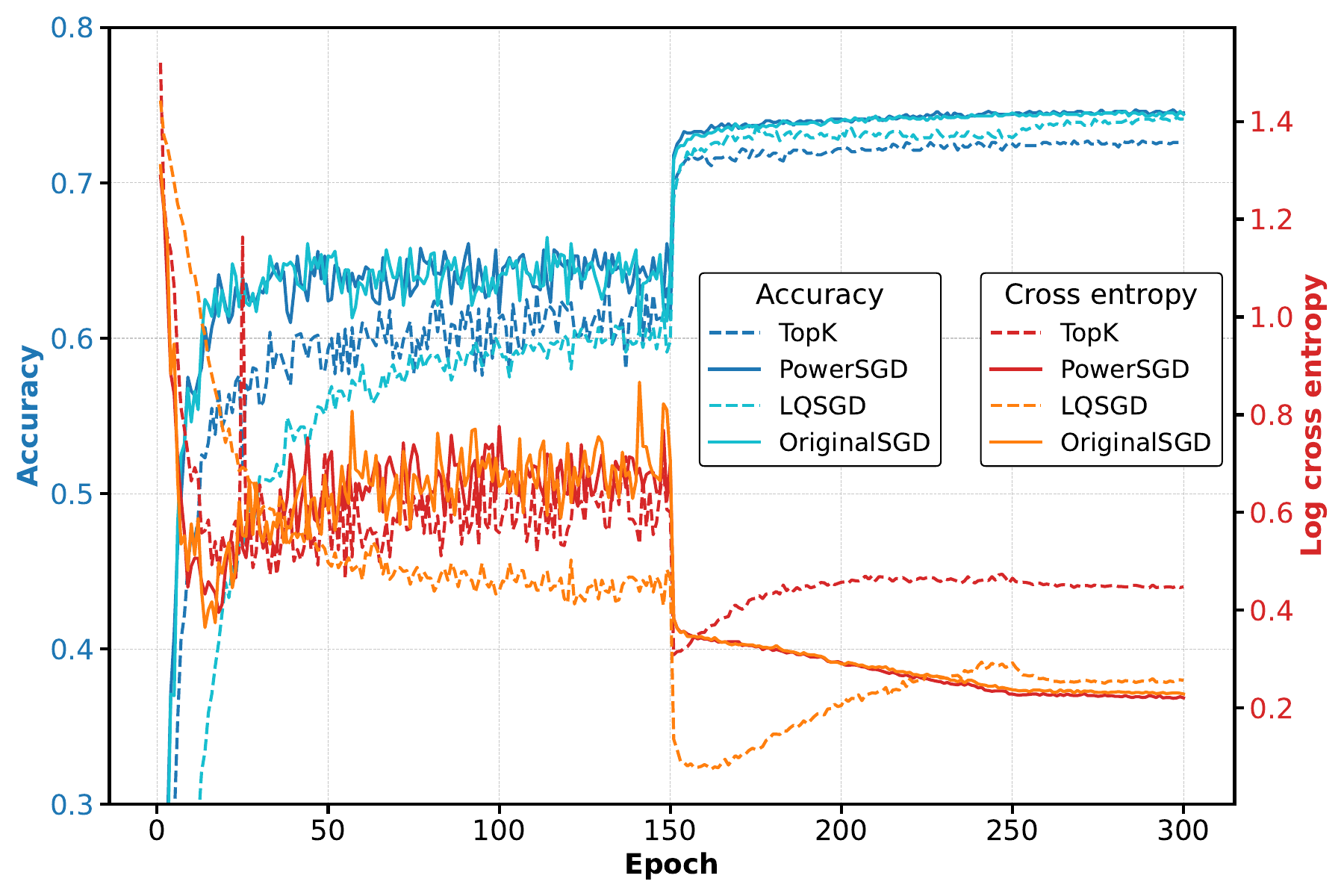}
        \caption{CIFAR-100, Rank 50}
        \label{fig:cifar100_rank50}
    \end{subfigure}
    
    \caption{Results for CIFAR-100 dataset with different compression ranks.}

    \label{fig:cifar100_selected_results}
\end{figure*}

\begin{figure*}[t]
    \centering
    \begin{subfigure}{0.4\textwidth}
        \centering
        \includegraphics[width=0.8\textwidth]{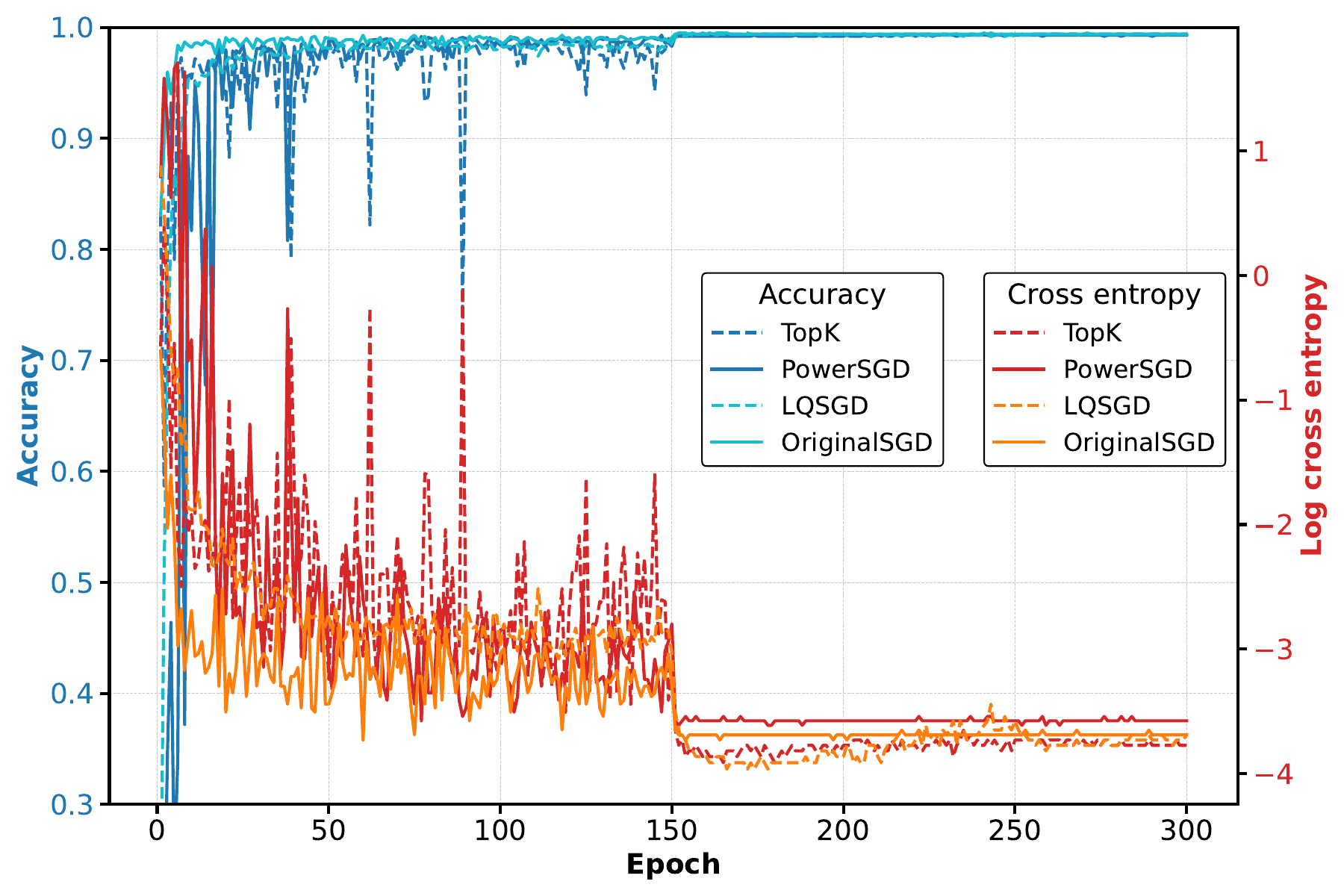}
        \caption{MNIST, Rank 1}
        \label{fig:mnist_rank1}
    \end{subfigure}
    \hfill
    \begin{subfigure}{0.4\textwidth}
        \centering
        \includegraphics[width=0.8\textwidth]{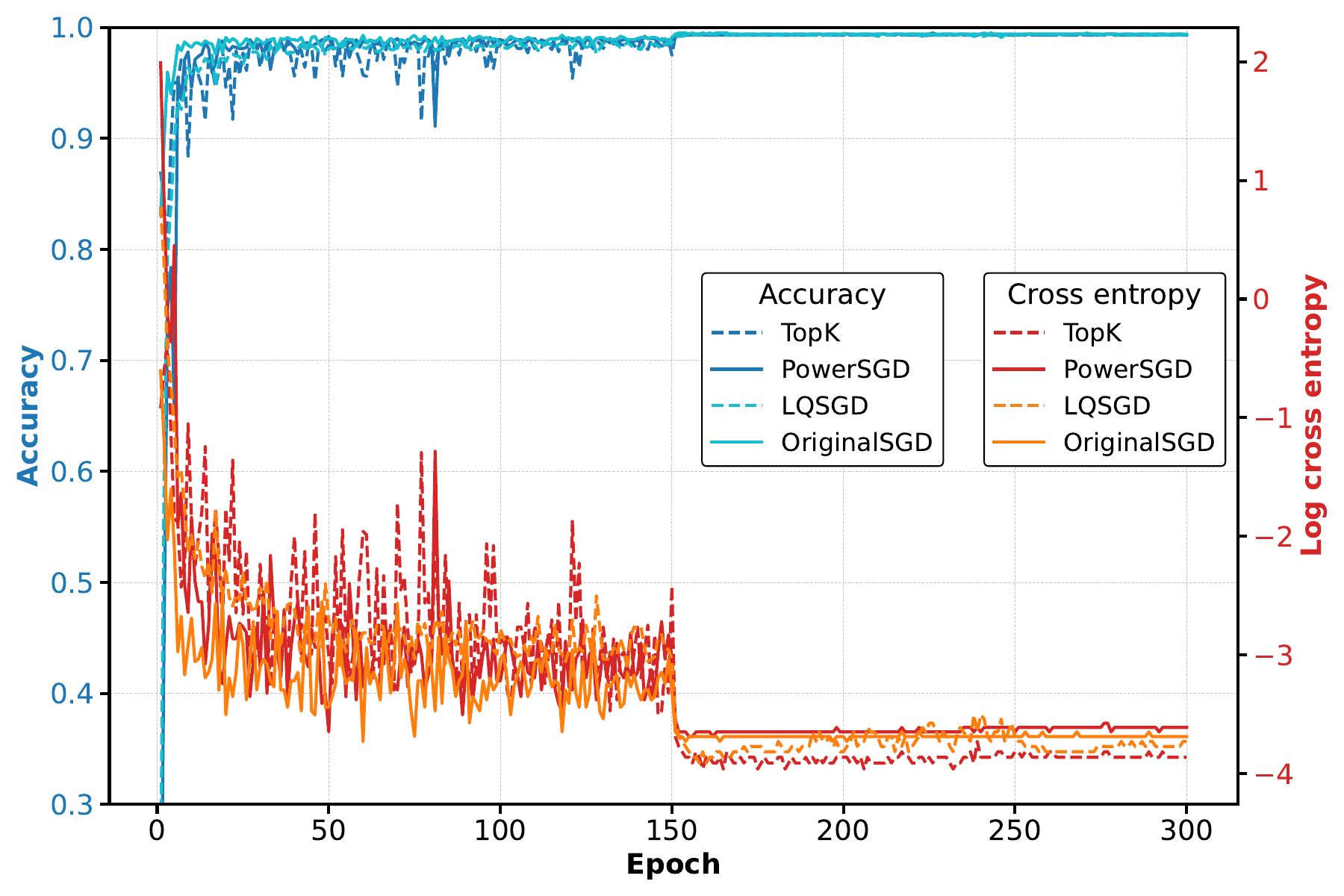}
        \caption{MNIST, Rank 2}
        \label{fig:mnist_rank2}
    \end{subfigure}
    \hfill
    \begin{subfigure}{0.4\textwidth}
        \centering
        \includegraphics[width=0.8\textwidth]{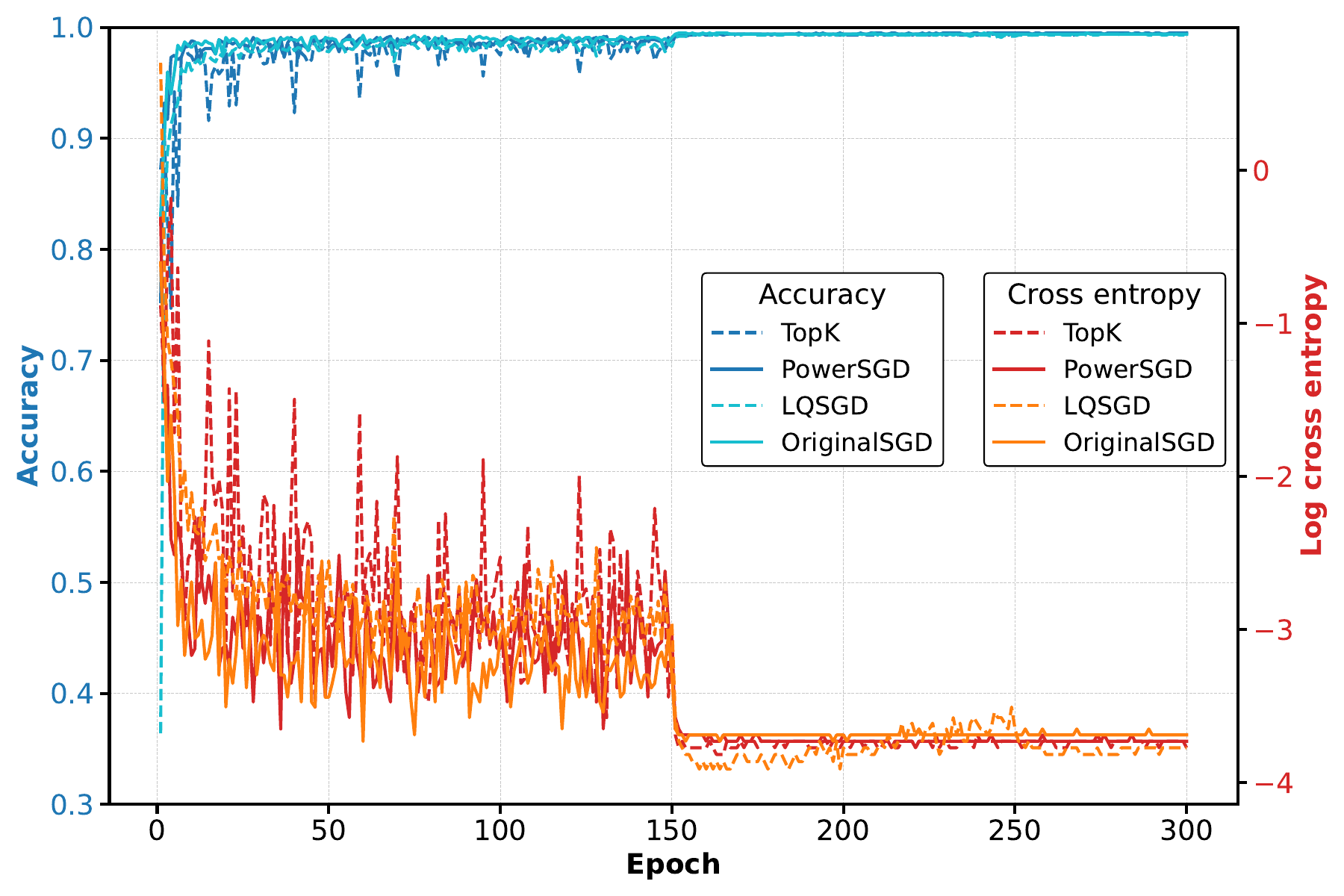}
        \caption{MNIST, Rank 4}
        \label{fig:mnist_rank4}
    \end{subfigure}
    \hfill
    \begin{subfigure}{0.4\textwidth}
        \centering
        \includegraphics[width=0.8\textwidth]{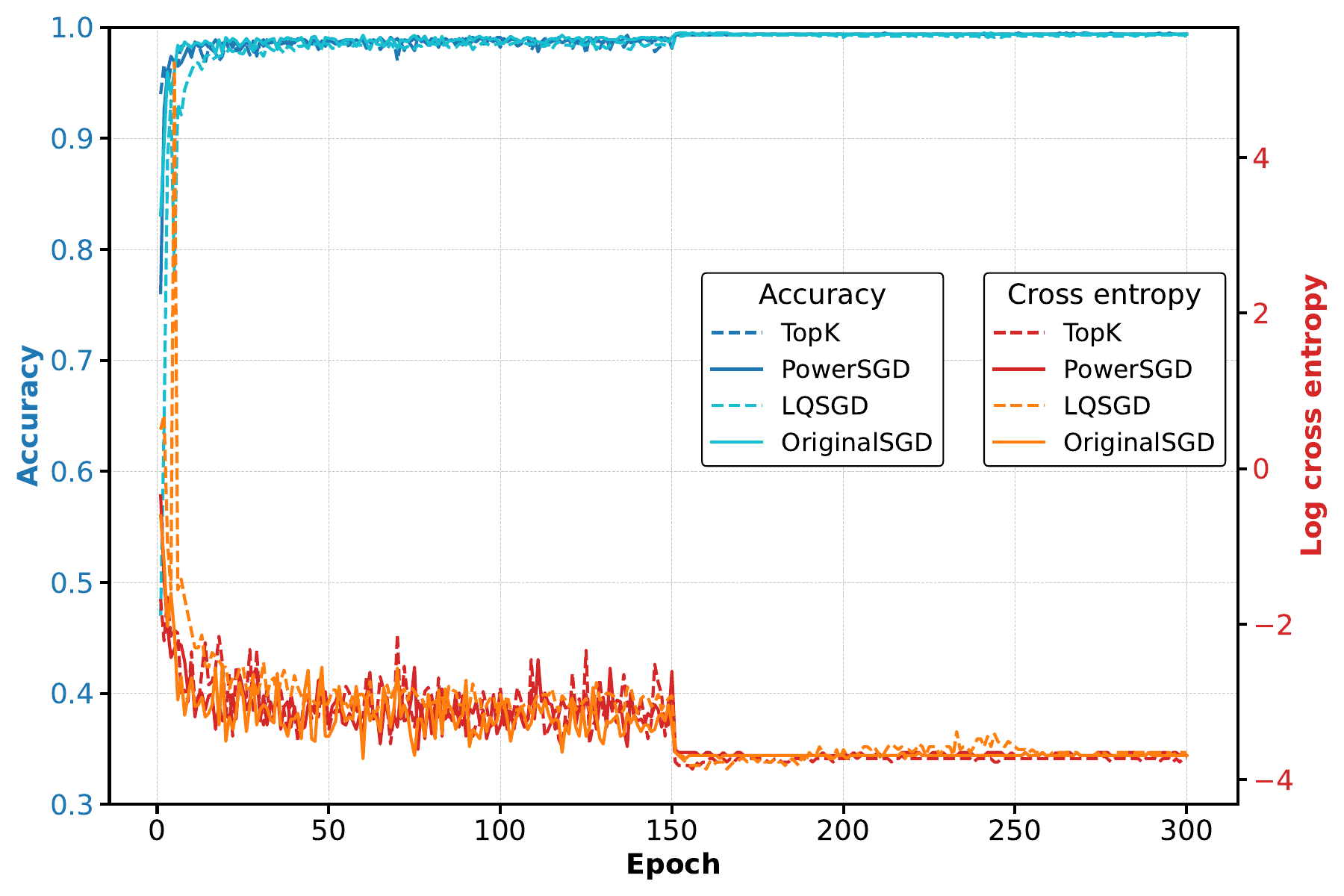}
        \caption{MNIST, Rank 30}
        \label{fig:mnist_rank30}
    \end{subfigure}
    \hfill
    \begin{subfigure}{0.4\textwidth}
        \centering
        \includegraphics[width=0.8\textwidth]{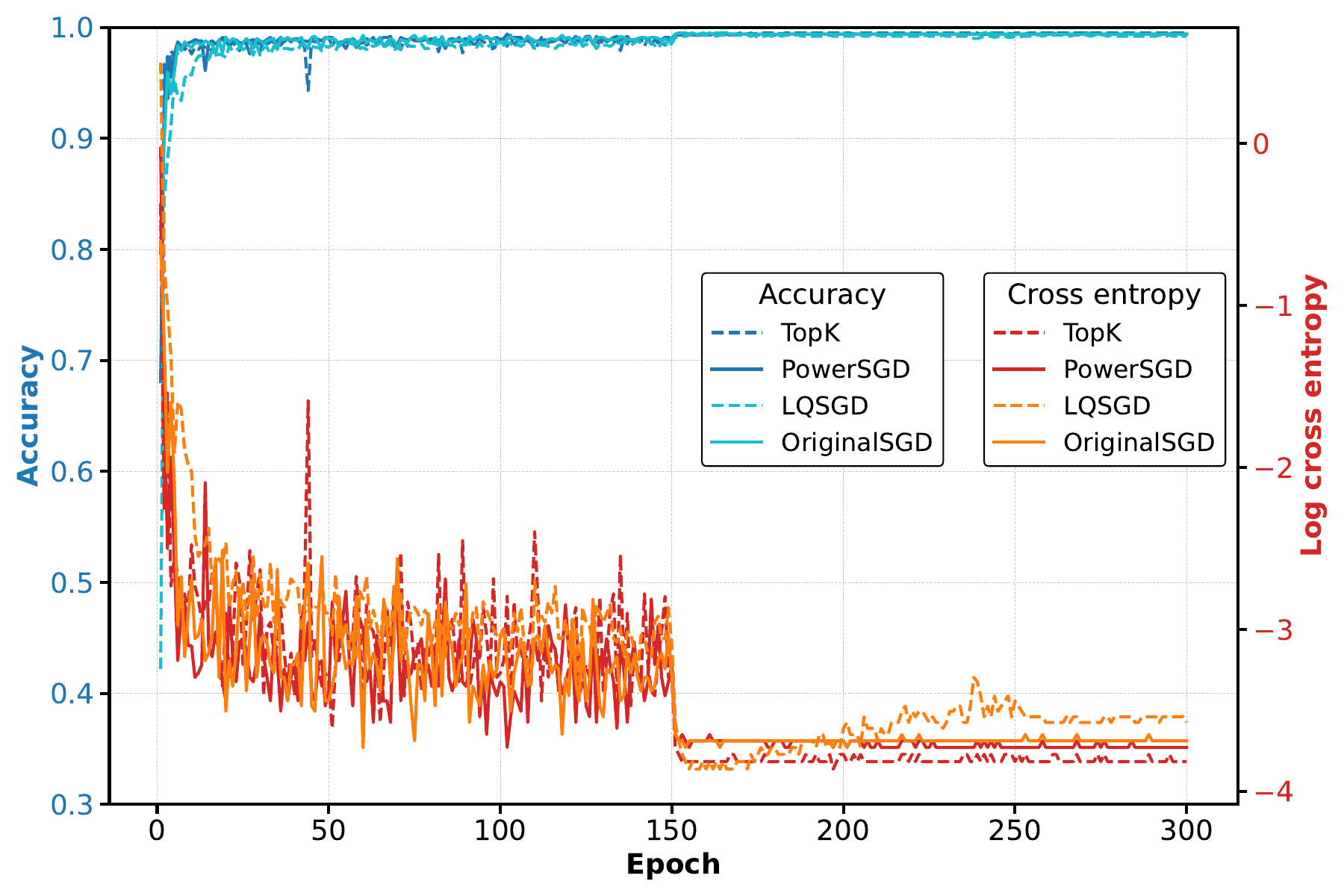}
        \caption{MNIST, Rank 50}
        \label{fig:mnist_rank50}
    \end{subfigure}
    
    \caption{Results for MNIST dataset with different compression ranks.}
    \caption*{\footnotesize $^\dagger$Note: Top K-SGD does not have a specific rank; the 'Rank' values are provided for consistency in compression rate comparisons.}
    \label{fig:mnist_selected_results}
\end{figure*}

\textbf{Overall Efficiency.} In large-scale distributed training, communication time often dominates, accounting for over 50\% of the total training time~\cite{anthony2024demystifying}. By significantly reducing communication volume, LQ-SGD can substantially lower total training time with negligible impact from the slightly increased computation time. Figures~\ref{fig:cifar10_selected_results}, \ref{fig:cifar100_selected_results}, and \ref{fig:mnist_selected_results} further demonstrate LQ-SGD's convergence behavior. It achieves convergence rates comparable to PowerSGD and TopK-SGD, stabilizing within 150 epochs across all datasets at different compression level.

\begin{figure}[ht]
    \centering
    \begin{subfigure}{0.4\textwidth}
        \includegraphics[width=\textwidth]{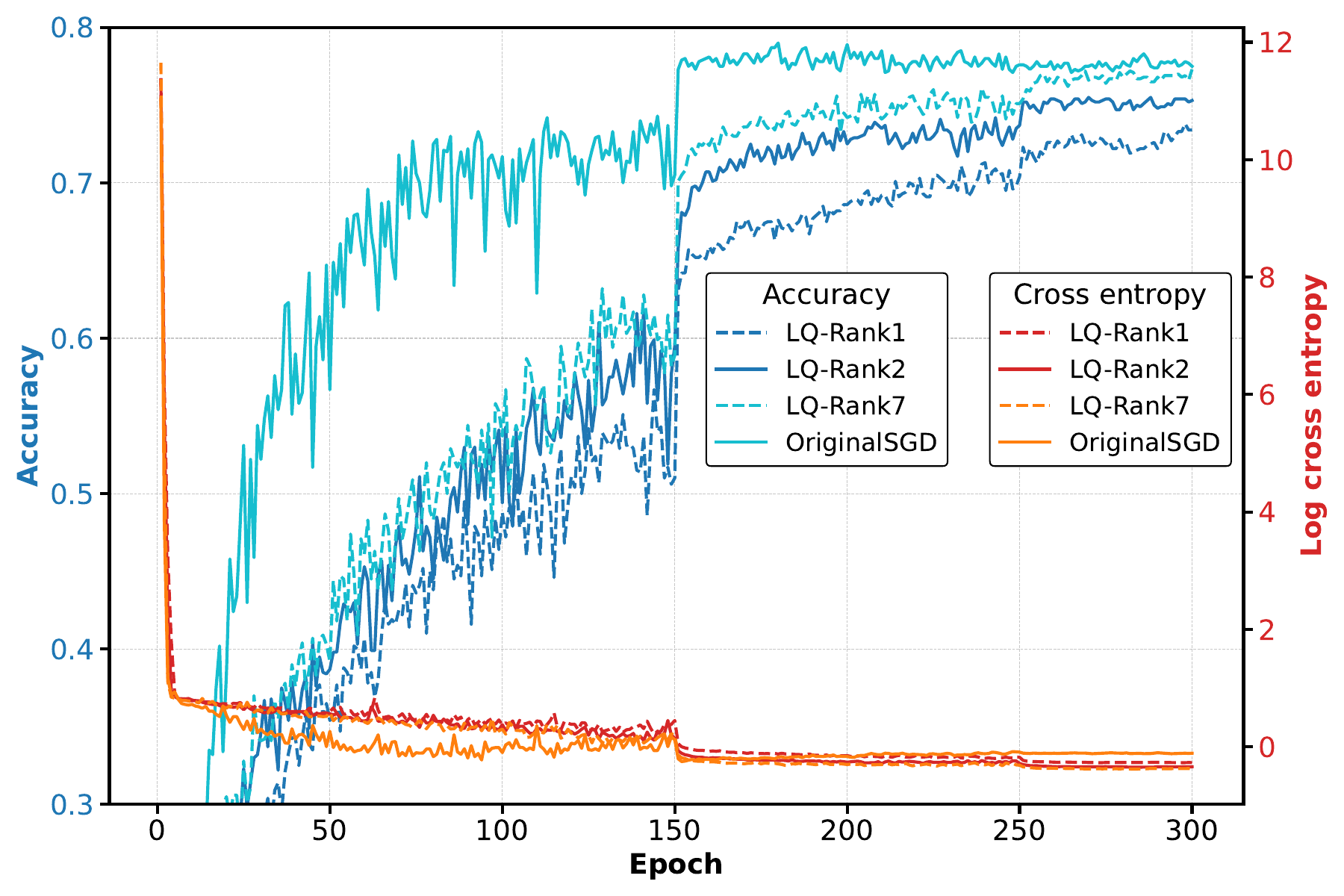}
        \caption{ImageNet}
        \label{fig:imagenet_rank7}
    \end{subfigure}
    \caption{ Demonstrates stable convergence up to 300 epochs.}
    \label{fig:imagenet_results}
\end{figure}

\text{\textit{\textbf{Performance on ImageNet}}}: Figure~\ref{fig:imagenet_results} presents the Top-1 accuracy and cross-entropy loss of LQ-SGD on ImageNet. With Rank 7, LQ-SGD matches the accuracy and convergence of OriginalSGD, reaching 75\% Top-1 accuracy after 300 epochs. Rank 2 maintains stable convergence with slightly reduced accuracy, while Rank 1 still achieves reasonable performance despite its aggressive compression.

These results confirm that LQ-SGD provides an excellent trade-off between communication cost and model accuracy, making it highly practical for bandwidth-constrained, large-scale distributed training.

\vspace{-2mm}

\begin{figure}[t]
\centering
\begin{subfigure}[b]{0.4\linewidth}
    \centering
    \includegraphics[width=\linewidth]{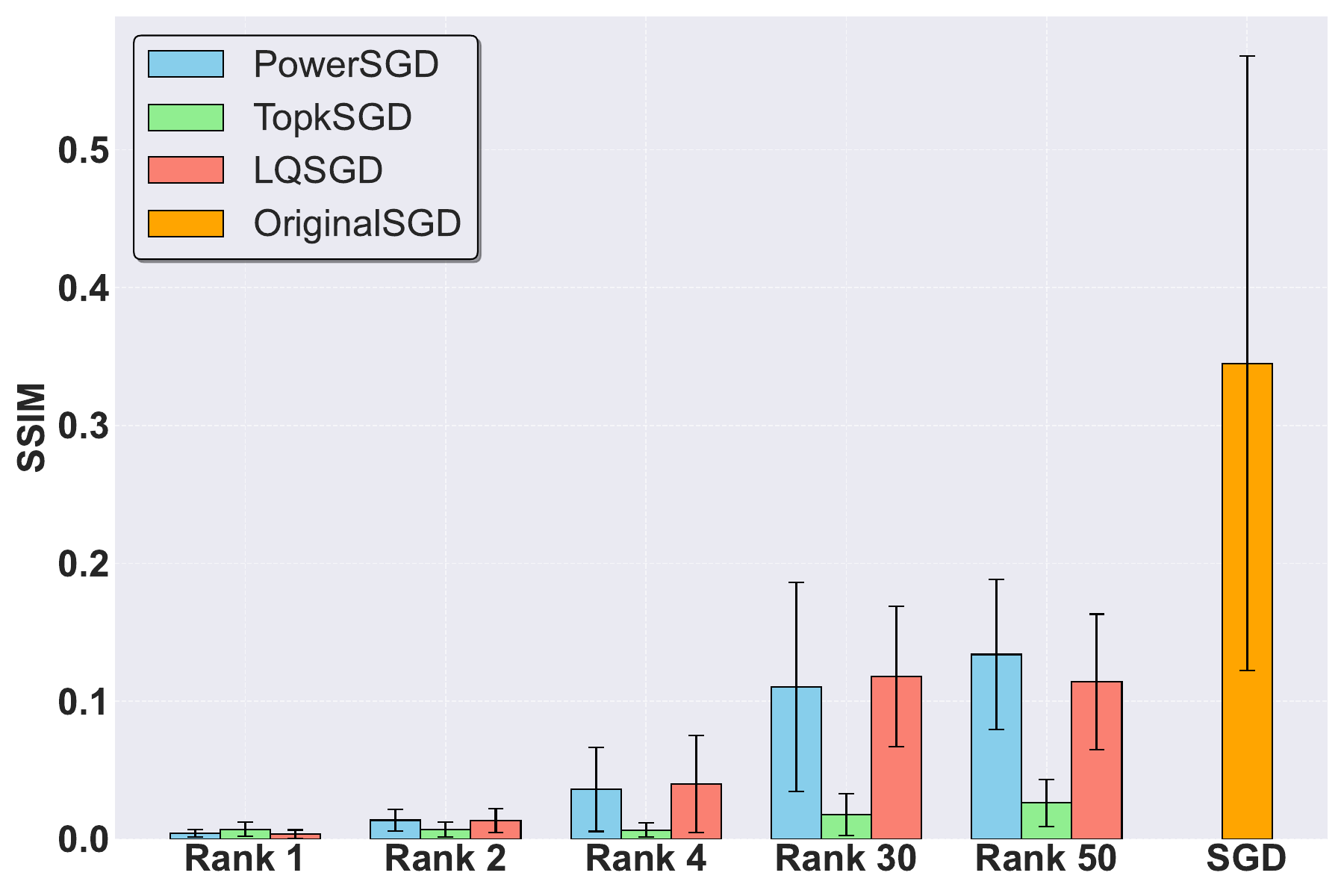}
    \caption{CIFAR-10}
    \label{fig:cifar10_ssim}
\end{subfigure}
\hfill
\begin{subfigure}[b]{0.4\linewidth}
    \centering
    \includegraphics[width=\linewidth]{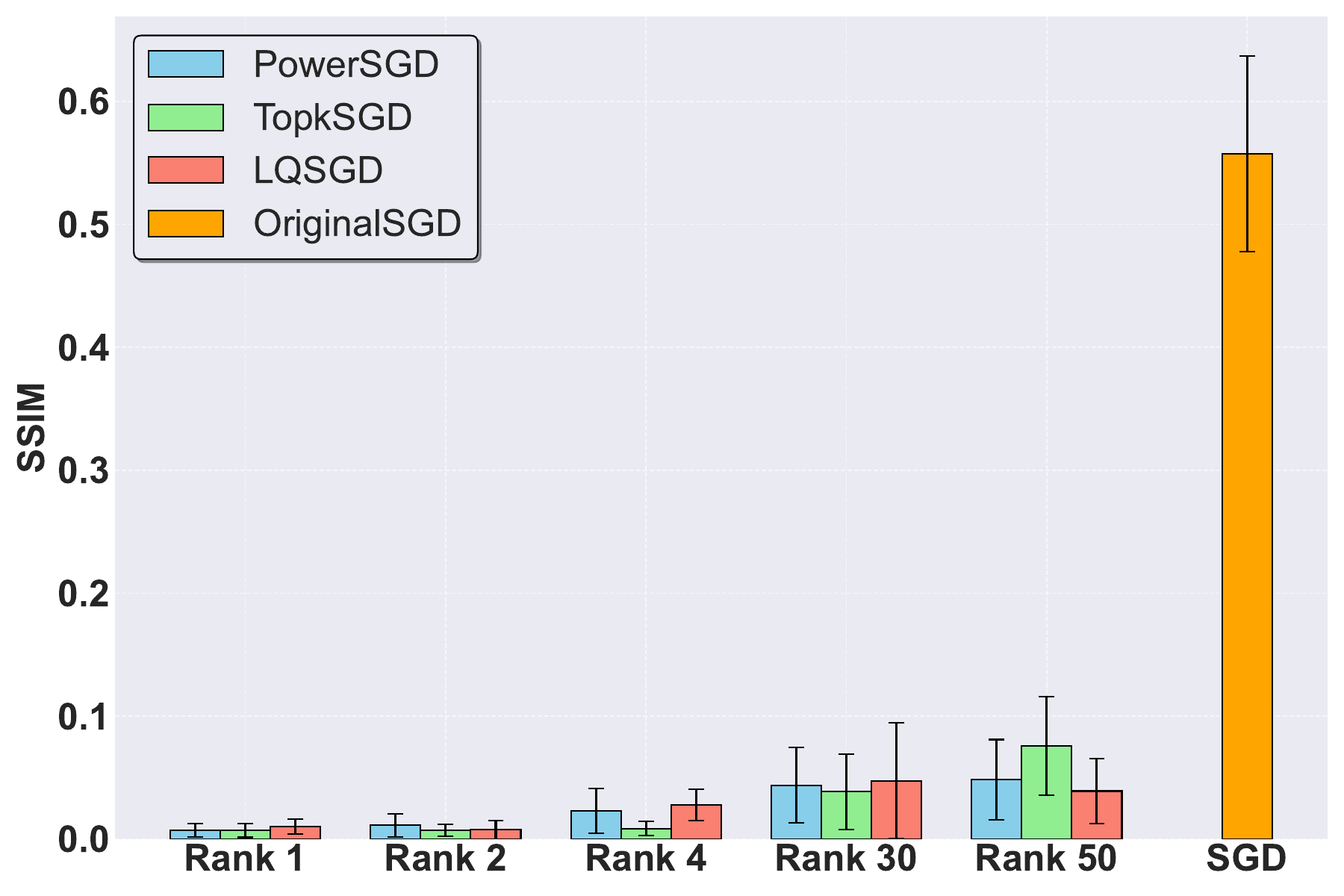}
    \caption{CIFAR-100}
    \label{fig:cifar100_ssim}
\end{subfigure}
\hfill
\begin{subfigure}[b]{0.4\linewidth}
    \centering
    \includegraphics[width=\linewidth]{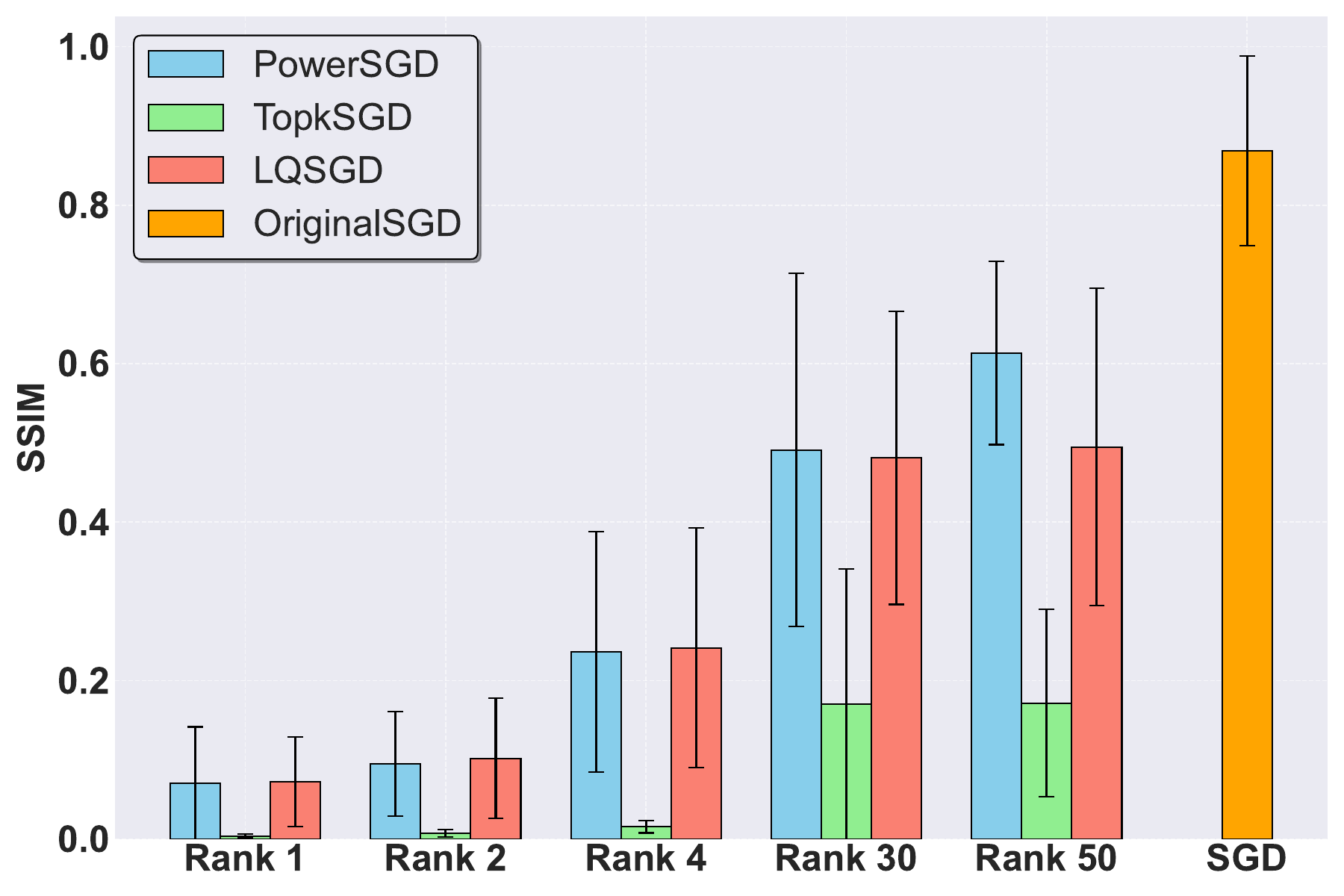}
    \caption{MNIST}
    \label{fig:mnist_ssim}
\end{subfigure}
\caption{SSIM scores under different compression ranks.}
\label{fig:ssim_combined}
\end{figure}

\subsection{Trustworthiness Evaluation: Resistance to Gradient Inversion Attacks}
We evaluate the privacy-preserving capabilities of LQ-SGD by conducting gradient GIA on CIFAR-10, CIFAR-100, and MNIST. To quantify the similarity between reconstructed and original images. Figures~\ref{fig:cifar10_ssim}--\ref{fig:mnist_ssim} present the SSIM results across different compression ranks for LQ-SGD, PowerSGD, TopK-SGD, and vanilla SGD.

Across all datasets, compression-based methods, including LQ-SGD, consistently yield lower SSIM scores compared to original SGD, indicating enhanced resistance to gradient inversion attacks. Among them, LQ-SGD achieves a favorable balance between privacy protection and model accuracy. Notably, while TopK-SGD obtains the lowest SSIM at high compression ratios, it often does so at the cost of substantial accuracy degradation. In contrast, LQ-SGD maintains strong model performance while offering meaningful privacy benefits.

These results confirm that gradient compression can effectively mitigate privacy leakage in distributed learning, and highlight LQ-SGD as a trustworthy and efficient choice for privacy-sensitive applications. This observation is in line with our earlier findings that gradient compression naturally enhances robustness against GIA \cite{li2024trustworthiness}.

\vspace{-3mm}

\section{Discussion}

LQ-SGD effectively balances communication efficiency with model performance and ensures trustworthiness. However, it has two main limitations: (1) our evaluation is limited to image classification tasks, and its scalability to diverse data has yet to be tested; (2) we have not evaluated LQ-SGD on very large architectures (e.g., transformer-based LLMs).

    \vspace{-3mm}

\section{Conclusion}

In this paper, we presented LQ-SGD and highlight it as a trustworthy and communication-efficient solution for distributed learning. In future work, we plan to extend LQ-SGD to diverse data sets and much larger network sizes.

    \vspace{-3mm}

\bibliography{references}  

\end{document}